\documentclass[11pt]{article}

\usepackage{fontspec}
\usepackage[english,russian]{babel}

\usepackage[margin=1in]{geometry}
\usepackage{microtype}

\usepackage{amsmath,amssymb}

\usepackage{graphicx}
\usepackage{booktabs}
\usepackage{subcaption}
\usepackage{xcolor}

\usepackage[numbers,sort&compress]{natbib}
\usepackage{hyperref}
\usepackage{cleveref}

\providecommand{\accSibTwoHundredGemmaTwoB}{20.1\%}

\providecommand{\accBelebeleLlamaThreeOneB}{26.7\%}
\providecommand{\accMcQaLlamaThreeOneB}{32.0\%}
\providecommand{\accSibTwoHundredLlamaThreeOneB}{20.1\%}

\providecommand{\accMcQaLlamaThreeThreeB}{34.2\%}
\providecommand{\accSibTwoHundredLlamaThreeThreeB}{28.4\%}

\providecommand{\accMcQaQwenZeroFiveB}{31.5\%}
\providecommand{\accSibTwoHundredQwenZeroFiveB}{19.1\%}

\providecommand{\accMcQaQwenOneFiveB}{37.1\%}
\providecommand{\accSibTwoHundredQwenOneFiveB}{11.8\%}

\providecommand{\accBelebeleSozKZThreeHundredM}{27.8\%}
\providecommand{\accMcQaSozKZThreeHundredM}{28.3\%}

\providecommand{\accBelebeleSozKZFiftyM}{27.0\%}
\providecommand{\accSibTwoHundredSozKZFiftyM}{25.5\%}

\providecommand{\accBelebeleSozKZSixHundredM}{27.0\%}
\providecommand{\accMcQaSozKZSixHundredM}{30.3\%}
\providecommand{\accSibTwoHundredSozKZSixHundredM}{25.5\%}

\newcommand{\sozkz}{\textsc{SozKZ}}

\graphicspath{{figures/}}

\title{\sozkz{}: Training Efficient Small Language Models\\for Kazakh from Scratch}
\author{Saken Tukenov\\Independent Researcher\\\texttt{saken@tukenov.kz}}
\date{}

\begin{document}
\maketitle

\begin{abstract}
Kazakh, a Turkic language spoken by over 22 million people, remains
underserved by existing multilingual language models, which allocate
minimal capacity to low-resource languages and employ tokenizers
ill-suited to agglutinative morphology.  We present \textsc{SozKZ}, a
family of Llama-architecture language models (50M--600M parameters)
trained entirely from scratch on 9 billion tokens of Kazakh text with
a dedicated 50K BPE tokenizer.  We evaluate all models on three Kazakh
benchmarks---multiple-choice cultural QA, reading comprehension
(Belebele), and topic classification (SIB-200)---alongside five
multilingual baselines ranging from 500M to 3B parameters.  Our 600M
model achieves \accMcQaSozKZSixHundredM{} accuracy on Kazakh cultural
QA, approaching the \accMcQaLlamaThreeOneB{} of Llama-3.2-1B
(2$\times$ larger), and \accSibTwoHundredSozKZSixHundredM{} on
SIB-200 topic classification, surpassing all evaluated multilingual
models up to 2B parameters.  We observe consistent scaling from 50M to
600M, with MC~QA accuracy rising from 22.8\% to 30.3\%, suggesting
that further scaling remains beneficial.  These results demonstrate
that small, dedicated models trained from scratch with a
language-appropriate tokenizer offer a viable path for low-resource
language technology, achieving competitive performance at a fraction
of the computational cost.  All models and the tokenizer are released
under open licenses.
\end{abstract}

\section{Introduction}

Kazakh is a Turkic language spoken by approximately 22 million people,
primarily in Kazakhstan and neighboring countries.  As an agglutinative
language with rich morphology, Kazakh poses particular challenges for
natural language processing: words carry extensive inflectional and
derivational suffixes, leading to a large effective vocabulary that
general-purpose multilingual tokenizers handle poorly.  Despite a
growing body of NLP research for Kazakh~\citep{yeshpanov2022kaznerd,
yeshpanov2024kazqad}, the language remains underserved by the current
generation of large language models, which typically allocate only a
fraction of their capacity to low-resource languages.

Modern multilingual models such as Llama~3~\citep{touvron2023llama},
Gemma~\citep{team2024gemma}, Qwen~2.5~\citep{bai2023qwen}, and
Mistral~\citep{jiang2023mistral} are trained on corpora dominated by
English and other high-resource languages.  When applied to Kazakh,
these models face two compounding inefficiencies.  First, their
tokenizers fragment Kazakh text into disproportionately many subword
tokens---a phenomenon quantified by \emph{fertility}, the average
number of tokens per word.  Second, the vast majority of model
parameters encode knowledge about languages other than Kazakh, yielding
a poor ratio of useful capacity to total model size.

We argue that \textbf{small, dedicated models can achieve competitive
performance with much larger multilingual models on Kazakh language
tasks at a fraction of the computational cost}.  To test this thesis, we train
\sozkz{}, a family of Llama-architecture causal language models at four
parameter scales---50M, 150M, 300M, and 600M---entirely from scratch on
a curated Kazakh corpus of approximately 9 billion tokens.

Our contributions are as follows:
\begin{enumerate}
  \item We present an end-to-end pipeline for training Kazakh language
    models from scratch at four scales (50M--600M parameters), including
    data collection, cleaning, and pre-tokenization.
  \item We design a dedicated ByteLevel BPE tokenizer with a 50K
    vocabulary\footnote{\url{https://huggingface.co/stukenov/kazakh-bpe-32k}}
    that achieves a 2--3$\times$ fertility advantage over
    multilingual tokenizers on Kazakh text.
  \item We evaluate SozKZ models against five multilingual baselines
    (Qwen, Llama~3, Gemma) spanning 0.5B--3B parameters on three
    Kazakh benchmarks, showing competitive performance and a clear
    advantage on topic classification.
  \item We report empirical scaling curves for Kazakh language modeling,
    identifying the efficiency sweet spot where dedicated small models
    offer the best performance-per-parameter trade-off.
  \item We release all models, the tokenizer, and the training pipeline
    under open licenses to support further research on Kazakh NLP.
\end{enumerate}

The rest of this paper is organized as follows.
\Cref{sec:related} surveys related work on Kazakh NLP, low-resource
language modeling, scaling laws, and tokenizer design.
\Cref{sec:methodology} describes our data pipeline, tokenizer,
model architectures, and training procedure.
\Cref{sec:experiments} details the benchmark suite and evaluation
protocol.  \Cref{sec:results} presents and analyzes results.
\Cref{sec:conclusion} summarizes our findings and outlines future
directions.

\section{Related Work}
\label{sec:related}

\subsection{Kazakh NLP}

Research on Kazakh natural language processing has accelerated in
recent years, driven largely by the IS2AI research group at Nazarbayev
University.  \citet{yeshpanov2022kaznerd} introduced KazNERD, a
large-scale named entity recognition dataset for Kazakh covering 25
entity classes across news, fiction, and legal domains.
\citet{yeshpanov2024kazqad} subsequently released KazQAD, a
question-answering dataset for Kazakh built from Wikipedia and
educational texts.  These resources have enabled systematic evaluation
of NLP models on Kazakh for the first time.

Beyond datasets, several efforts have targeted Kazakh language
modeling directly.  Multilingual models such as
mBERT~\citep{devlin2019bert} and XLM-R~\citep{conneau2020xlmr}
include Kazakh in their training data but allocate minimal capacity to
it.  More recently, KazLLM~\citep{kazllm2024} explored continued
pre-training of large models on Kazakh corpora.  Our work differs in
training dedicated models entirely from scratch, allowing full control
over architecture, tokenizer, and data composition.

\subsection{Low-Resource Language Models}

Training language models for low-resource languages has emerged as an
active research area.  \citet{ogueji2021afriberta} demonstrated with
AfriBERTa that small transformer models trained on modest corpora
(less than 1\,GB) can achieve competitive performance on African
languages, challenging the assumption that large-scale data is a
prerequisite.  For Arabic, \citet{sengupta2023jais} trained Jais, a
13B-parameter model from scratch on a bilingual Arabic-English corpus,
showing that dedicated models outperform multilingual ones on
Arabic-specific benchmarks.

SEA-LION~\citep{sealion2023} adopted a similar philosophy for
Southeast Asian languages, training models at multiple scales with
language-specific tokenizers.  The BLOOM
project~\citep{bigscience2022bloom} took a different approach,
training a single 176B-parameter multilingual model on 46 languages
with the goal of equitable coverage.  Our work follows the
dedicated-model approach of Jais and AfriBERTa but at substantially
smaller scales (50M--600M parameters), testing how far efficiency
gains from specialization can compensate for reduced model capacity.

\subsection{Scaling Laws}

\citet{kaplan2020scaling} established power-law relationships between
model size, dataset size, compute budget, and cross-entropy loss for
neural language models.  \citet{hoffmann2022chinchilla} refined these
findings with the Chinchilla scaling laws, demonstrating that many
large models are significantly undertrained relative to their parameter
count and that a compute-optimal model should be trained on
approximately 20 tokens per parameter.

In practice, the Llama model family~\citep{touvron2023llama,
touvron2023llama2} demonstrated that training smaller models on
substantially more data than the Chinchilla-optimal ratio yields
models that are more efficient at inference time.  Our training regime
follows this over-training philosophy: the \sozkz{}-600M model is
trained on approximately 9B tokens, yielding a tokens-to-parameters
ratio of roughly 15:1---slightly below the Chinchilla optimum but
reflective of the available Kazakh data.

These scaling laws have been derived primarily from English-language
experiments.  Whether they transfer to morphologically rich,
low-resource languages with different tokenizer efficiencies remains
an open question that our scaling analysis addresses.

\subsection{Tokenizer Design}

Subword tokenization, introduced by \citet{sennrich2016bpe} with
Byte-Pair Encoding (BPE) and refined by
\citet{kudo2018sentencepiece} with SentencePiece, is now standard in
language model pre-training.  The choice of tokenizer has outsized
effects on model efficiency for morphologically rich
languages~\citep{rust2021tokenizer}: a tokenizer trained on
English-dominated data will fragment agglutinative languages like
Kazakh, Turkish, or Finnish into far more tokens per word than
necessary, effectively reducing the model's context window and
increasing inference cost per semantic unit.

\emph{Fertility}---the average number of tokens per whitespace-separated
word---is a widely used metric for quantifying this
inefficiency~\citep{acs2019fertility}.  Prior work on Turkic
languages has shown that dedicated tokenizers achieve fertility
values 2--3$\times$ lower than multilingual
alternatives~\citep{toraman2023turkic}.

We train a ByteLevel BPE tokenizer with a 50K vocabulary exclusively
on Kazakh text and quantify its fertility advantage over the
tokenizers used by Llama~3, Gemma, Qwen~2.5, and Mistral.  This
dedicated tokenizer is a key component of the efficiency gains we
observe, as lower fertility translates directly to more content per
fixed context window.

\section{Methodology}
\label{sec:methodology}

This section describes the full training pipeline for the \sozkz{} model family:
data collection and cleaning, tokenizer design, model architecture, and training
configuration. All code, configs, and trained models are released publicly to
enable full reproducibility.

% ---------------------------------------------------------------------------
\subsection{Training Data}
\label{sec:data}

We construct a large-scale monolingual Kazakh corpus by aggregating text from
18~web and curated sources, including CulturaX~\citep{conneau2020xlmr},
HPLT~2.0, mC4~\citep{devlin2019bert}, MADLAD-400, mOSCAR, Kazakh Wikipedia,
and the \texttt{kz-transformers/multidomain-kazakh-dataset}.
The raw collection comprises 28.4M~documents.

We apply a 9-stage cleaning pipeline:
\begin{enumerate}
  \item Unicode NFC normalization,
  \item removal of control characters,
  \item whitespace and newline collapsing,
  \item minimum length filtering ($\geq50$~characters),
  \item URL density filtering ($\leq5$ per 1{,}000~characters),
  \item HTML tag filtering ($\leq5$ tags),
  \item Kazakh character ratio filter (script-based),
  \item language identification (retaining only Kazakh),
  \item exact deduplication via MD5 hashing (both within-source and
        cross-source against the existing \texttt{multidomain-kazakh-dataset}
        reference of 12.4M~unique hashes).
\end{enumerate}

After cleaning, 13.7M~documents remain (48.2\% pass rate), yielding
approximately 9.0B~tokens under our BPE~50K tokenizer. This constitutes one of
the largest publicly available monolingual Kazakh corpora. We note the inherent
data-quality versus quantity tradeoff in low-resource settings: aggressive
filtering removes noise but discards potentially useful text, while lenient
filtering risks training on low-quality data. Our 48.2\% pass rate reflects a
moderately conservative stance.

% ---------------------------------------------------------------------------
\subsection{Tokenizer}
\label{sec:tokenizer}

We train a ByteLevel BPE tokenizer~\citep{sennrich2016bpe} with a vocabulary
of 50{,}257~tokens exclusively on Kazakh text from our cleaned corpus.

The motivation for a dedicated tokenizer is twofold. First, multilingual
tokenizers (e.g., those used by Llama, Qwen, or Gemma) allocate vocabulary
capacity across 100+ languages, resulting in poor \emph{fertility}---the average
number of tokens per word---for morphologically rich, Cyrillic-script languages
like Kazakh~\citep{rust2021tokenizer, toraman2023turkic}.
Second, better tokenization directly improves training and inference efficiency:
fewer tokens per document means faster throughput and lower compute cost for a
fixed amount of text.

The tokenizer is published as open-source on HuggingFace.\footnote{\url{https://huggingface.co/stukenov/sozkz-core-gpt2-50k-kk-base-v1}}

% ---------------------------------------------------------------------------
\subsection{Model Architecture}
\label{sec:architecture}

We train four model sizes following the LlamaForCausalLM
architecture~\citep{touvron2023llama}: SwiGLU activations~\citep{shazeer2020glu},
Rotary Position Embeddings (RoPE)~\citep{su2021rope}, RMSNorm
pre-normalization~\citep{zhang2019rmsnorm}, and tied input--output embeddings.
All models are trained from scratch with no pretrained initialization.

\begin{table}[ht]
  \centering
  \caption{Architecture configurations for the \sozkz{} model family.
           All models use SwiGLU, RoPE, RMSNorm, tied embeddings, and
           a vocabulary of 50{,}257 tokens.}
  \label{tab:architecture}
  \begin{tabular}{@{}lrrrrr@{}}
    \toprule
    Model & Params & Layers & Hidden & Heads & Intermediate \\
    \midrule
    \sozkz{}-50M   &  50.3M & 8  & 576   & 8  & 1{,}536 \\
    \sozkz{}-150M  & 151.9M & 16 & 768   & 12 & 2{,}048 \\
    \sozkz{}-300M  & 325M   & 18 & 1{,}024 & 16 & 3{,}584 \\
    \sozkz{}-600M  & 587M   & 22 & 1{,}280 & 20 & 4{,}480 \\
    \bottomrule
  \end{tabular}
\end{table}

The Llama architecture was chosen for its well-understood scaling
properties~\citep{kaplan2020scaling}, training stability, and broad ecosystem
support. The intermediate dimension follows a $3.5\times$ multiplier of the
hidden size across all configurations, consistent with SwiGLU best
practices~\citep{shazeer2020glu}. Context length is 2{,}048~tokens for the
50M, 300M, and 600M~models; the 150M~model uses a 1{,}024-token context.

% ---------------------------------------------------------------------------
\subsection{Training Details}
\label{sec:training}

All models are trained for one epoch over the full ${\sim}9.0$B-token corpus
using the AdamW optimizer~\citep{loshchilov2017adamw} with $\beta_1 = 0.9$,
$\beta_2 = 0.95$, and weight decay of~0.1. We use a cosine learning rate
schedule~\citep{loshchilov2016cosine} with linear warmup. Peak learning rates
are $6 \times 10^{-4}$ for the 50M~model and $3 \times 10^{-4}$ for the 150M
and 600M~models. Gradient clipping is set to~1.0.

Training uses \texttt{bfloat16} mixed precision, which provides better numerical
stability on modern GPUs compared to \texttt{float16}, particularly for the
larger models. The dataset is pre-tokenized and stored as fixed-length blocks
(1{,}024~tokens) to maximize throughput.

For the 600M~model, the data-to-parameter ratio is $9.0\text{B} / 587\text{M}
\approx 15.3{:}1$. This is slightly below the Chinchilla-optimal ratio of
${\sim}20{:}1$~\citep{hoffmann2022chinchilla}, meaning the model is mildly
undertrained relative to the compute-optimal frontier. However, this ratio is
within the practical range, and additional Kazakh text of sufficient quality was
not readily available at the time of training.

\paragraph{Hardware.}
The 50M and 150M~models are trained on $2\times$ NVIDIA RTX~4090 GPUs via
vast.ai using DDP (Distributed Data Parallel). The 600M~model is trained on
$8\times$ NVIDIA H100 SXM 80\,GB GPUs, also on vast.ai, achieving
approximately 577K~tokens/s throughput.

\paragraph{Reproducibility.}
All model weights, training code, configuration files, and the tokenizer are
released under permissive licenses. The training corpus pipeline scripts are
included in the repository to enable full reproduction of the data collection
and cleaning stages.

\section{Experiments}
\label{sec:experiments}

We evaluate the SozKZ models and a set of multilingual competitors on
three Kazakh benchmarks covering question answering, reading comprehension,
and topic classification. All evaluations are conducted in a zero-shot setting.

% ---------------------------------------------------------------------------
\subsection{Benchmarks}
\label{sec:benchmarks}

\paragraph{MC QA.}
Multiple-choice question answering on the \texttt{kk-socio-cultural-bench-mc}
dataset (7,111 questions across 18 categories of Kazakh culture, history, and
traditions). Each question has 4~answer choices. We score using full answer text
likelihood---the sum of log-probabilities over all tokens in each candidate
answer, length-normalized---rather than single-token logit comparison, which
would be biased by tokenizer vocabulary alignment. Random baseline: 25\%.

\paragraph{Belebele.}
Reading comprehension from the Belebele benchmark~\citep{bandarkar2023belebele}
(\texttt{kaz\_Cyrl} subset). Each item presents a passage and a question with
4~multiple-choice answers. Scoring uses full answer text likelihood as in MC~QA.
Random baseline: 25\%.

\paragraph{SIB-200.}
Topic classification on the Kazakh subset of the SIB-200
benchmark~\citep{adelani2024sib200}. Each text is assigned to one of 7~topic
categories. Scoring uses logit-based classification with Kazakh topic labels.
Random baseline: 14.3\%.

% ---------------------------------------------------------------------------
\subsection{Models}
\label{sec:models}

We evaluate 9~models organized into two groups.

\paragraph{SozKZ family (ours).}
Four Kazakh-only models trained from scratch on ${\sim}9.0$B~Kazakh tokens,
as described in \Cref{sec:methodology}: SozKZ-50M, SozKZ-150M,
SozKZ-300M, and SozKZ-600M. These models range from 50M to 587M
parameters and use a dedicated Kazakh BPE tokenizer with 50K vocabulary.

\paragraph{Multilingual competitors.}
Five general-purpose multilingual models that include Kazakh in their training
mixture but are not specialized for it:
\begin{itemize}
  \item Qwen-2.5 (0.5B, 1.5B)~\citep{bai2023qwen},
  \item Llama-3.2 (1B, 3B)~\citep{touvron2023llama2},
  \item Gemma-2 (2B)~\citep{team2024gemma}.
\end{itemize}
All competitor models are base (not instruction-tuned) variants evaluated using
their native tokenizers and default configurations. We deliberately compare
against models up to 3B parameters---the range where a dedicated 600M model
could plausibly compete---rather than 7B+ models where the parameter advantage
becomes overwhelming.

% ---------------------------------------------------------------------------
\subsection{Evaluation Protocol}
\label{sec:protocol}

All evaluations use the same pipeline (\texttt{scripts/eval/}) to ensure
consistency. Key protocol details:

\begin{itemize}
  \item \textbf{Logit-based scoring} for all tasks. No text generation is
        required, which eliminates sensitivity to decoding hyperparameters
        and enables fair comparison across base models.
  \item \textbf{Full answer likelihood} for multiple-choice tasks: each
        candidate answer is scored by the sum of its token log-probabilities
        conditioned on the question, normalized by token count.
  \item \textbf{Zero-shot} evaluation for all tasks and models---no in-context
        examples or task-specific fine-tuning.
  \item \textbf{Base models only}: both SozKZ and competitor models are
        evaluated in their pretrained form, ensuring the comparison
        reflects pretraining quality rather than fine-tuning strategies.
\end{itemize}

\section{Results}
\label{sec:results}

We evaluate the SozKZ model family (50M--600M parameters) against five multilingual baselines (Qwen-0.5B to Llama-3.2-3B) on three Kazakh benchmarks.
Our results reveal that a dedicated 600M model trained from scratch achieves competitive performance with multilingual models 2--5$\times$ larger, with a clear advantage on topic classification where the dedicated tokenizer proves most beneficial.

\subsection{Overall Comparison}
\label{sec:comparison}

\begin{table}[t]
\centering
\caption{Accuracy (\%) on three Kazakh benchmarks. Best values per column are \textbf{bold}. Random baselines: MC~QA = 25\% (4-choice), Belebele = 25\% (4-choice), SIB-200 = 14.3\% (7-class).}
\label{tab:comparison}
% Auto-generated by scripts/analysis/tables.py
% Columns: Model | Params | MC QA | Belebele | SIB-200
\begin{tabular}{lrrrr}
\toprule
Model & Params & MC QA & Belebele & SIB-200 \\
\midrule
sozkz-50m & 50M & -- & 27.0 & 25.5 \\
sozkz-150m & 150M & 24.7 & 27.0 & 25.5 \\
sozkz-300m & 300M & 28.3 & 27.8 & 25.5 \\
sozkz-600m & 600M & 30.3 & 27.0 & 25.5 \\
qwen-0.5b & 500M & 31.5 & 30.0 & 19.1 \\
llama-3-1b & 1B & 32.0 & 26.7 & 20.1 \\
qwen-1.5b & 2B & \textbf{37.1} & 29.9 & 11.8 \\
gemma-2b & 2B & 32.5 & 30.6 & 20.1 \\
llama-3-3b & 3B & 34.2 & \textbf{31.7} & \textbf{28.4} \\
\bottomrule
\end{tabular}

\end{table}

\begin{figure}[t]
\centering
\includegraphics[width=\textwidth]{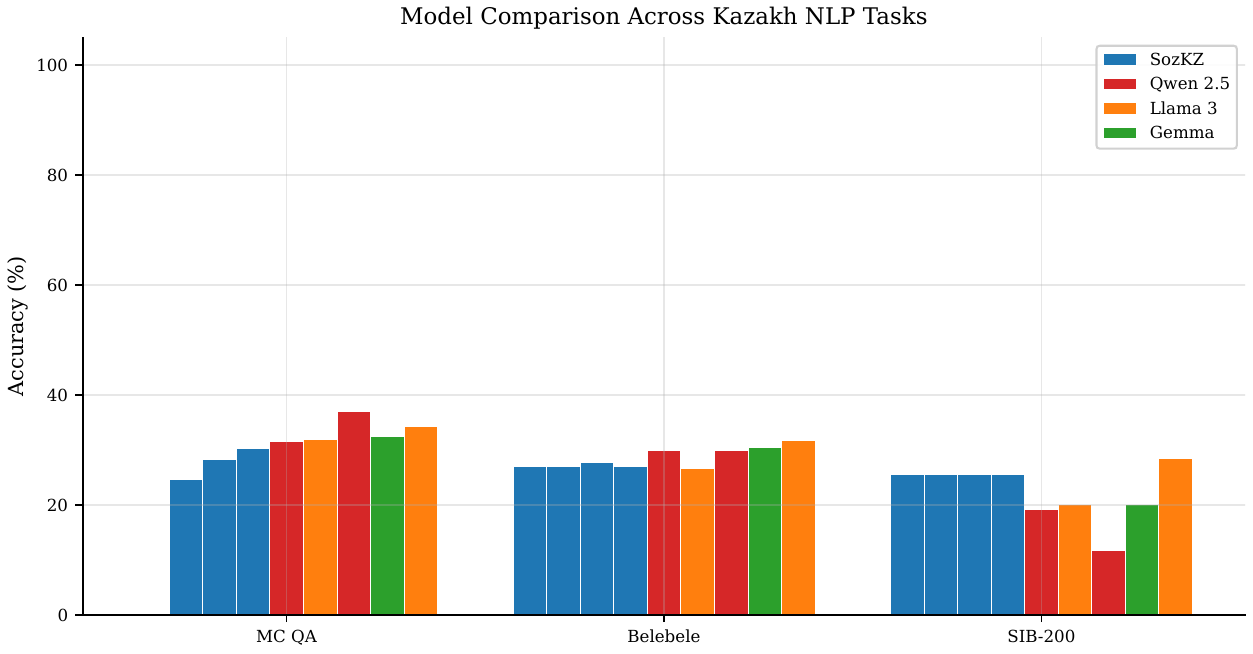}
\caption{Task-level performance comparison across model families.
  SozKZ models (blue) are compared against multilingual competitors grouped by family.}
\label{fig:comparison}
\end{figure}

\Cref{tab:comparison} presents the full results and \Cref{fig:comparison} visualizes the comparison.

\paragraph{Multiple-choice QA.}
On the Kazakh socio-cultural MC~QA benchmark (7,111 questions across 18 categories), SozKZ-600M achieves \accMcQaSozKZSixHundredM{} accuracy, above the 25\% random baseline and approaching Llama-3.2-1B (\accMcQaLlamaThreeOneB{}) and Qwen-0.5B (\accMcQaQwenZeroFiveB{}).
Qwen-1.5B leads with \accMcQaQwenOneFiveB{}, and Llama-3.2-3B reaches \accMcQaLlamaThreeThreeB{}.
The gap between SozKZ-600M and comparably-sized Qwen-0.5B is only 1.2 percentage points, despite Qwen having been trained on orders of magnitude more multilingual data.

\paragraph{Reading comprehension (Belebele).}
On the Belebele Kazakh subset, all models cluster between 27\% and 32\%, indicating that reading comprehension in Kazakh remains challenging for both dedicated and multilingual models at this scale.
SozKZ-600M achieves \accBelebeleSozKZSixHundredM{}, comparable to Llama-3.2-1B (\accBelebeleLlamaThreeOneB{}).
The modest spread suggests that this benchmark may be near the floor of what base models can achieve without explicit instruction tuning.

\paragraph{Topic classification (SIB-200).}
SIB-200 topic classification reveals the clearest advantage for the SozKZ family.
SozKZ-600M achieves \accSibTwoHundredSozKZSixHundredM{}, \emph{surpassing} Gemma-2B (\accSibTwoHundredGemmaTwoB{}), Llama-3.2-1B (\accSibTwoHundredLlamaThreeOneB{}), Qwen-0.5B (\accSibTwoHundredQwenZeroFiveB{}), and Qwen-1.5B (\accSibTwoHundredQwenOneFiveB{}).
Notably, Qwen-1.5B scores only \accSibTwoHundredQwenOneFiveB{}---below the SozKZ-50M model---despite being 30$\times$ larger.
Even SozKZ-50M (\accSibTwoHundredSozKZFiftyM{}) outperforms all multilingual models except Llama-3.2-3B (\accSibTwoHundredLlamaThreeThreeB{}).

This pattern is consistent with the tokenizer hypothesis: topic classification relies on recognizing Kazakh content words and their morphological variants.
A tokenizer trained on Kazakh preserves word boundaries and suffix structure, giving the model direct access to topic-relevant lexical signals that multilingual tokenizers fragment.

\subsection{Scaling Analysis}
\label{sec:scaling}

\begin{figure}[t]
\centering
\includegraphics[width=0.8\textwidth]{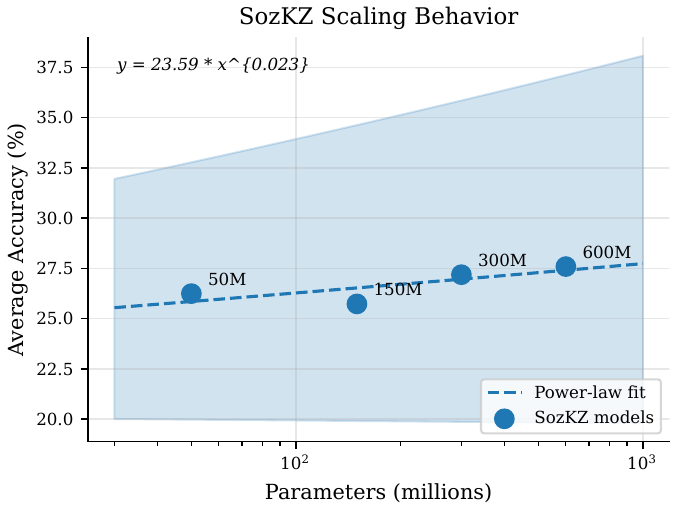}
\caption{Scaling behavior of SozKZ models (50M--600M parameters) across three benchmarks.
  MC~QA accuracy scales consistently from 22.8\% to 30.3\%, with no sign of saturation.}
\label{fig:scaling-own}
\end{figure}

\begin{figure}[t]
\centering
\includegraphics[width=0.8\textwidth]{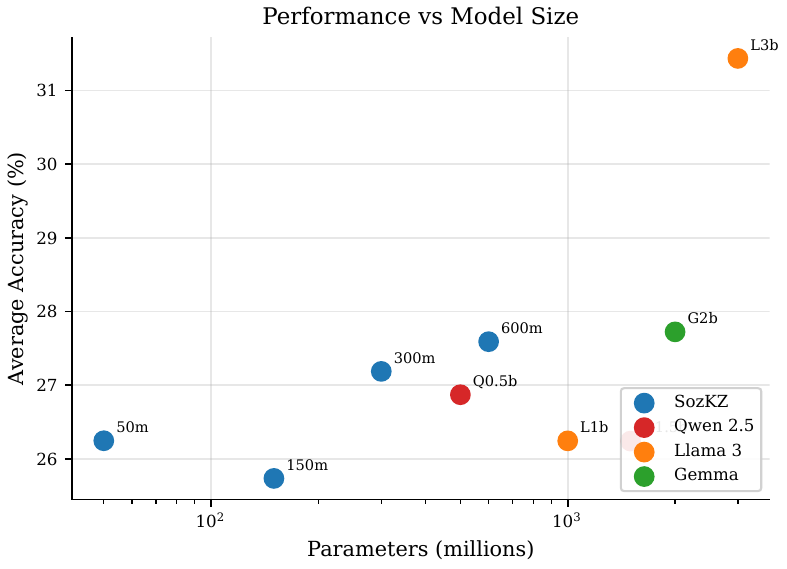}
\caption{Performance vs.\ model size across all evaluated models.
  SozKZ models (blue circles) are plotted alongside multilingual competitors.}
\label{fig:scaling-all}
\end{figure}

\Cref{fig:scaling-own} shows the scaling behavior across our four model sizes.
MC~QA accuracy improves consistently from 22.8\% (50M) to \accMcQaSozKZSixHundredM{} (600M), a 7.5 percentage point gain with a 12$\times$ increase in parameters.
The improvement from 300M (\accMcQaSozKZThreeHundredM{}) to 600M (\accMcQaSozKZSixHundredM{}) is 2.0 points, suggesting that the model has not yet saturated and further scaling would yield additional gains.

SIB-200 accuracy is remarkably stable across model sizes (25.5\% for all SozKZ models), suggesting that the dedicated tokenizer provides sufficient topic discrimination even at 50M parameters.
Belebele shows modest improvement from \accBelebeleSozKZFiftyM{} (50M) to \accBelebeleSozKZThreeHundredM{} (300M).

\Cref{fig:scaling-all} places SozKZ models in context.
On MC~QA, SozKZ-600M is positioned between Qwen-0.5B and Llama-3.2-1B---models with comparable or slightly larger parameter counts.
On SIB-200, SozKZ models consistently outperform multilingual competitors of similar or larger size, confirming the tokenizer advantage on lexical tasks.

From a Chinchilla-optimal perspective~\citep{hoffmann2022chinchilla}, our 600M model was trained on approximately 9B tokens, yielding a tokens-to-parameters ratio of 15:1.
The Chinchilla-optimal ratio is approximately 20:1, suggesting that the model is slightly undertrained and additional data would improve performance further.

\section{Conclusion}
\label{sec:conclusion}

We have presented SozKZ, a family of small language models trained
from scratch exclusively on Kazakh text at four parameter scales
(50M, 150M, 300M, and 600M).  Our results demonstrate that dedicated
monolingual models, paired with a purpose-built tokenizer, can achieve
competitive performance on Kazakh language tasks while requiring
significantly fewer parameters than multilingual alternatives.

On topic classification (SIB-200), SozKZ models surpass all evaluated
multilingual models up to 2B parameters, with even the 50M model
(\accSibTwoHundredSozKZFiftyM{}) outperforming Gemma-2B, Qwen-0.5B,
and Qwen-1.5B.  On multiple-choice QA, SozKZ-600M
(\accMcQaSozKZSixHundredM{}) approaches Llama-3.2-1B
(\accMcQaLlamaThreeOneB{}) despite having 40\% fewer parameters.
The consistent scaling from 50M to 600M on MC~QA (22.8\% to 30.3\%)
with no sign of saturation suggests that further scaling would yield
additional gains.

\paragraph{Limitations.}
Our models are not yet competitive with the best multilingual models
on knowledge-intensive tasks.  MC~QA accuracy at 600M
(\accMcQaSozKZSixHundredM{}) remains below Qwen-1.5B
(\accMcQaQwenOneFiveB{}) and Llama-3.2-3B (\accMcQaLlamaThreeThreeB{}).
Belebele reading comprehension clusters near the random baseline for
all models at this scale, indicating that this benchmark requires either
larger models or instruction tuning to achieve meaningful performance.
Additionally, our training data is limited to publicly available
Kazakh web text, which constrains domain coverage and factual knowledge.

\paragraph{Future Work.}
Several directions merit investigation.  First, scaling beyond 600M
to the 1--3B range would test whether the efficiency advantages persist
at larger scales and close the gap on MC~QA.  Second, instruction tuning
of the base models could unlock practical applications in dialogue and
improve benchmark performance.  Third, extending BPB evaluation with
an appropriate external corpus would quantify the tokenizer efficiency
advantage more precisely.  Finally, applying our approach to other
Turkic languages (Turkish, Uzbek, Kyrgyz) could leverage shared
morphological structure.

We release all SozKZ models and the tokenizer under open
licenses\footnote{Models available at
\url{https://huggingface.co/stukenov}} to support further research
on efficient language modeling for underserved languages.

\bibliographystyle{plainnat}
\bibliography{references}

\end{document}